\documentclass{article}

\usepackage{microtype}
\usepackage{graphicx}
\usepackage{subfigure}
\usepackage{booktabs} 

\usepackage{hyperref}

\usepackage[accepted]{icml2023}

\usepackage{amsmath}
\usepackage{amssymb}
\usepackage{mathtools}
\usepackage{amsthm}

\usepackage[capitalize,noabbrev]{cleveref}

\theoremstyle{plain}

\theoremstyle{definition}

\theoremstyle{remark}

\usepackage[textsize=tiny]{todonotes}

\begin{document}

\twocolumn[
\icmltitle{A Survey of Vision Transformers in Autonomous Driving: \\
           Current Trends and Future Directions}



\icmlsetsymbol{equal}{*}

\begin{icmlauthorlist}
    \icmlauthor{Quoc-Vinh Lai-Dang}{yyy}
\end{icmlauthorlist}

\icmlaffiliation{yyy}{Cho Chun Shik Graduate School of Mobility, Daejeon, Korea}

\icmlcorrespondingauthor{Quoc-Vinh Lai-Dang}{ldqvinh@kaist.ac.kr}


\vskip 0.3in
]



\printAffiliationsAndNotice{}  

\begin{abstract}
This survey explores the adaptation of visual transformer models in Autonomous Driving, a transition inspired by their success in Natural Language Processing. Surpassing traditional Recurrent Neural Networks in tasks like sequential image processing and outperforming Convolutional Neural Networks in global context capture, as evidenced in complex scene recognition, Transformers are gaining traction in computer vision. These capabilities are crucial in Autonomous Driving for real-time, dynamic visual scene processing. Our survey provides a comprehensive overview of Vision Transformer applications in Autonomous Driving, focusing on foundational concepts such as self-attention, multi-head attention, and encoder-decoder architecture. We cover applications in object detection, segmentation, pedestrian detection, lane detection, and more, comparing their architectural merits and limitations. The survey concludes with future research directions, highlighting the growing role of Vision Transformers in Autonomous Driving.

\icmlkeywords{Autonomous Driving, Vision Transformers, Machine Learning}
\end{abstract}

\section{Introduction}
Transformers \cite{vaswani2017attention} have revolutionized Natural Language Processing (NLP), with models like BERT, GPT, and T5 setting new standards in language understanding \cite{Alaparthi20, Radford18, Raffel20}. Their impact extends beyond NLP, as the Computer Vision (CV) community adopts Transformers for visual data processing. This shift from traditional Convolutional Neural Networks (CNNs) and Recurrent Neural Networks (RNNs) to Transformers in CV signifies their growing influence, with early implementations in image recognition and object detection \cite{Dosovitskiy20, carion2020end, Zhu20, kim} showing promising outcomes.

In Autonomous Driving (AD), Transformers are transforming a range of critical tasks, including object detection \cite{Mao23}, lane detection \cite{Han22}, and segmentation \cite{Ando23, Cakir22}, and can be combined with reinforcement learning \cite{Seo, Vecchietti} to execute complex path finding. They excel in processing spatial and temporal data, outperforming traditional CNNs and RNNs in complex functions like scene graph \cite{Liu22} generation and tracking \cite{Zhang23}. The self-attention mechanism of Transformers provides a more comprehensive understanding of dynamic driving environments, essential for the safe navigation of autonomous vehicles.

This survey offers an extensive overview of Vision Transformers in AD, exploring their development, taxonomy, and varied applications. Starting with foundational aspects of Transformer architecture, the paper progresses to examine their roles in AD, highlighting improvements in 3D and 2D perception tasks. Concluding with future research directions, it emphasizes the potential in advancing AD, aiming to inspire further exploration and application in this field.

\section{Exploring the Transformer: Structural and Functional Insights}
\subsection{The Transformer Architecture: A Structural Overview}
The Transformer architecture, a groundbreaking innovation by \cite{vaswani2017attention}, marks a departure from traditional recurrent layers by utilizing Attention mechanisms for sequence processing. It comprises two primary components: the Encoder and the Decoder. The Encoder processes input embeddings through Multi-head Attention and Feed-Forward Networks, both enhanced by Layer Normalization and residual connections. The Decoder, similar in structure to the Encoder, also focuses on the Encoder output, producing the final output sequence. Positional encodings are crucial in this architecture, as they imbue the model with the ability to recognize sequence order, a critical feature since Transformers do not inherently discern word order. This functionality is crucial for grasping language contexts, making positional encodings a fundamental component of the Transformer design. In the following, we describe each component of the Transformer in detail.

\begin{figure}[ht]
\vskip 0.1in
\begin{center}
\centerline{\includegraphics[width=0.5\columnwidth]{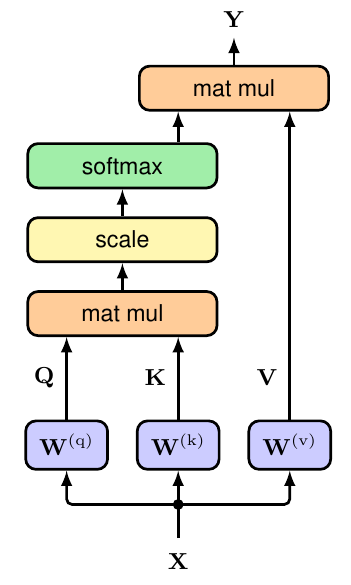}}
\caption{Self-attention process.}
\label{fig:self}
\end{center}
\vskip -0.4in
\end{figure}

\subsection{Self-Attention Mechanism: The Heart of Transformers}
Central to the Transformer model is the Self-Attention mechanism (\cref{fig:self}), which assesses how various segments of the input sequence are related to one another. In this process, each input element is converted into three vectors: queries (\textbf{q}), keys (\textbf{k}), and values (\textbf{v}), typically of dimension \(d = 512\), and compiled into matrices \(Q\), \(K\), and \(V\). The attention function then calculates interaction scores through a dot product between queries and keys, followed by normalization (dividing by \(\sqrt{d}\)) to stabilize training. These scores are converted into probabilities via a softmax function, indicating the degree of attention each element warrants. The final output (\(Y\)) is computed as:
\begin{equation}
    Y = \text{softmax}\left(\frac{Q \cdot K^T}{\sqrt{d}}\right) V. \label{eq:1}
\end{equation}
This is a weighted sum of the value vectors, encapsulating the context of the entire sequence. Additionally, the Encoder-Decoder Attention mechanism allows the decoder to concentrate on pertinent segments of the input sequence, informed by its present state and the output from the encoder. This mechanism, coupled with Positional encodings that add unique position information to input embeddings, ensures a comprehensive understanding of sequence ordering.

\subsection{Multi-Head Attention: Enhancing Dimensional Analysis}

\begin{figure}[ht]
\vskip 0.1in
\begin{center}
\centerline{\includegraphics[width=0.9\columnwidth]{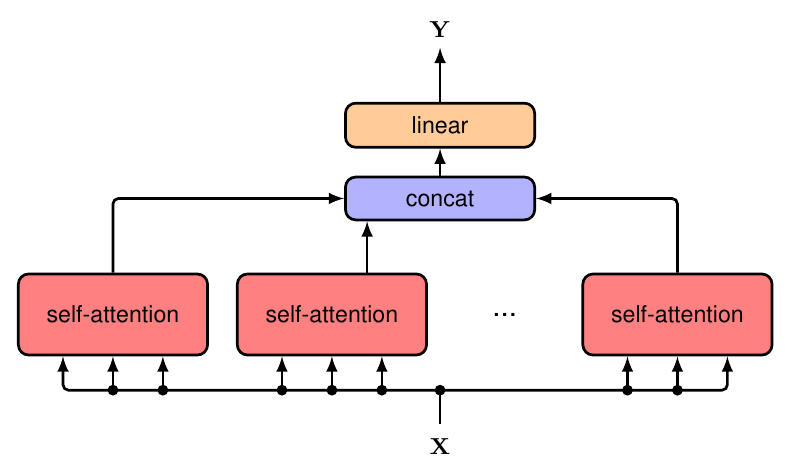}}
\caption{Multi-Head attention process.}
\label{fig:multi}
\end{center}
\vskip -0.3in
\end{figure}

The Multi-head Attention mechanism (\cref{fig:multi}) enhances the ability to analyze various dimensions of the input data. Initially, the input vector is partitioned into three distinct sets for each head: the query set \(Q'\), the key set \(K'\), and the value set \(V'\), with each subset having a dimension of \(\frac{d}{h}\). These sets are composed of smaller vectors—specifically, \(\textit{h}\) vectors per set, each with a dimension of 64 when \(d\) is 512. These vectors are then grouped to form the matrices \(Q'\), \(K'\), and \(V'\)for the subsequent attention calculations. The Multi-Head Attention process is formalized as follows:
\begin{equation}
    \text{MultiHead}(Q', K', V') = \text{Concat}(head_1, \ldots, head_h)W^O, \label{eq:2}
\end{equation}
where each \(\text{head}_i\) is defined as \(Y\). In this context,  \(Q'\), \(K'\) and \(V'\) represent the collective matrices formed by the concatenation of their respective vectors, and \(W^O\) is a matrix of learned weights that combines the individual output of attention heads into a single output vector.

\subsection{Other Core Mechanics of Transformer Models}
The Feed-Forward Network (FFN) is a crucial component of Transformer models, positioned after Self-Attention computations in each unit. It consists of a two-stage linear operation with a nonlinear activation function, typically the Gaussian Error Linear Unit (GELU)\cite{Hendrycks16}. This is mathematically represented as: 
\begin{equation}
    \text{FFN}(X) = W_2 \sigma(W_1 X), \label{eq:3}
\end{equation}
where \(W_1\) and \(W_2\) are matrices of learnable parameters, and \(\sigma\) represents the nonlinear function. The role of FFN is to augment the capability to process complex data patterns, with its intermediate layer usually housing about 2048 units.

Skip connections are integral to each layer of Transformer models, enhancing information flow and addressing vanishing gradient issues. These connections add the input directly to the output of sub-layers:
\begin{equation}
    \text{LayerNorm}(X+Z(X)), \label{eq:4}
\end{equation}
where \(X\) is the input and \(Z(X)\) is the output. Skip connections, combined with Layer Normalization \cite{ba2016layer}, ensure stable learning. Some variants use Pre-Layer Normalization \cite{xiong2020layer, takase2022layer, shleifer2021normformer} for optimization, applying normalization before each sub-layer.

The output layer in Transformers is vital for translating vector sequences into interpretable outputs. It involves linearly mapping vectors to a logits space matching the vocabulary size, followed by a softmax function that converts logits to a probability distribution. This layer is key to transforming processed data into final, understandable results, crucial in various data processing tasks.

Transformers in Autonomous Driving function as advanced feature extractors, differing from CNNs by integrating information across larger visual fields for global scene understanding. Their capability to process data in parallel offers significant computational efficiency, essential for real-time processing in autonomous vehicles. The global perspective and efficiency make the Transformer highly advantageous for Autonomous Driving technology, enhancing system capabilities.

\section{Vision Transformers in Autonomous Driving}

\begin{figure*}[ht]
\vskip 0.1in
\begin{center}
\centerline{\includegraphics[width=0.7\textwidth]{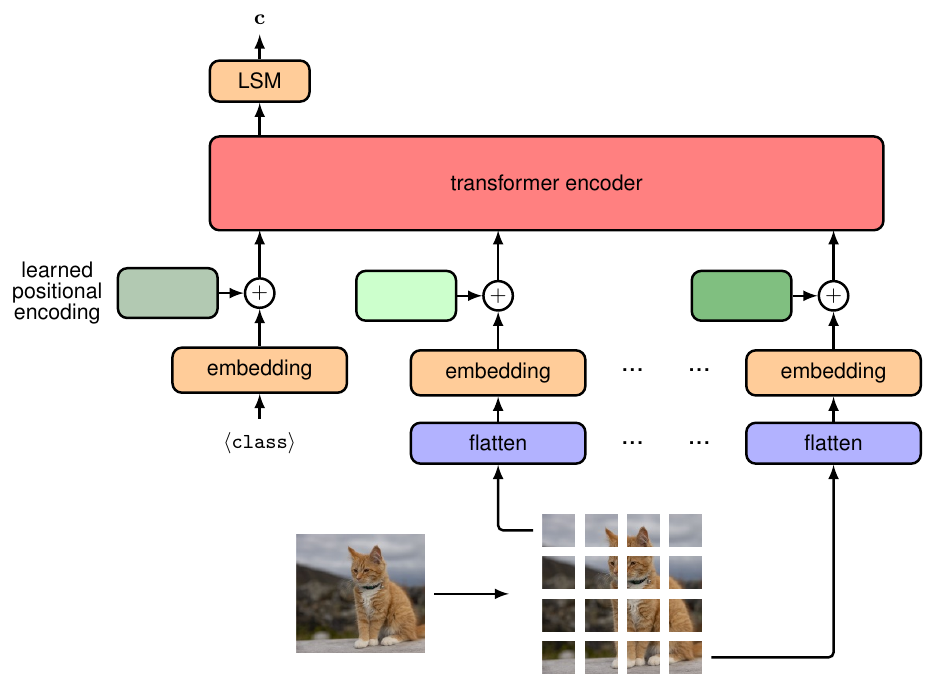}}
\caption{Vision Transformer Architecture.}
\label{fig:vit}
\end{center}
\vskip -0.3in
\end{figure*}

Building on the foundational concepts of vanilla Transformers primarily used in NLP, this section ventures into the dynamic world of Vision Transformers (ViTs) and their impactful role in AD. ViTs have significantly evolved, showcasing their versatility and effectiveness in vehicular technologies. The upcoming subsections will detail how ViTs are employed across various dimensions of autonomous driving. We start by exploring their involvement in 3D tasks, including essential functions like object detection, tracking, and 3D segmentation, which are fundamental for comprehensive environmental perception. The narrative then transitions to 2D tasks, highlighting their capabilities in lane detection, sophisticated segmentation, and high-definition map creation — all crucial for interpreting two-dimensional spatial data. Finally, we delve into other pivotal roles of ViTs, such as trajectory and behavior prediction and their integration within end-to-end Autonomous Driving systems. This journey through the applications of ViTs in Autonomous Driving not only demonstrates their adaptability but also emphasizes their growing importance in enhancing the capabilities of autonomous vehicles.

\subsection{The Rise of Vision Transformers}
The ViT \cite{Dosovitskiy20} (\cref{fig:vit}) has brought a paradigm shift in image processing within Autonomous Driving, replacing conventional convolutional layers with Self-Attention layers. This transformative approach segments images into distinct patches for analysis using a Transformer encoder, comprised of Self-Attention and Feed-Forward layers. This enables focused analysis on essential image segments, substantially improving perception in driving scenarios. For larger images, ViT adopts a hybrid model, combining convolutional and Self-Attention layers. This innovative strategy is crucial for efficiently processing complex visual data, a key requirement for the sophisticated decision-making necessary in autonomous vehicles.

Advancing the concepts introduced by ViT, the Swin-Transformer \cite{liu2021swin} presents a novel hierarchical structure, specifically designed for image processing in Autonomous Driving systems. It effectively addresses the scalability challenges of Vision Transformers, primarily due to the high computational demands of Self-Attention mechanisms. The introduction of shifted windows in the Swin-Transformer facilitates efficient attention to adjacent patches without overlap, significantly reducing computational load and enabling the processing of larger images. Additionally, its unique tokenization method, which segments images into fixed-size patches and groups them hierarchically, maintains critical spatial information and captures both local and global scene contexts. The Swin-Transformer's proficiency in processing image features has led to its widespread use in various Autonomous Driving perception models, such as BEVFusion \cite{liang2022bevfusion, liu2023bevfusion} and BEVerse \cite{zhang2022beverse}, highlighting its impact in advancing autonomous driving technology.

\subsection{3D Perception Tasks}
The application of Vision Transformer models has led to significant progress in 3D and general perception tasks within autonomous driving. Initial models such as DETR \cite{carion2020end} adopted an innovative method to object detection by framing it as a set prediction issue, employing pre-defined boxes and utilizing the Hungarian algorithm to predict sets of objects. This methodology was further refined in Deformable DETR (\cite{Zhu20}, which incorporated deformable attention for improved query clarity and faster convergence. DETR3D \cite{wang2022detr3d} extended these principles to 3D object detection, transforming LiDAR data into 3D voxel representations. Additionally, Vision Transformers like FUTR \cite{gong2022future} and FUTR3D \cite{chen2023futr3d} have broadened their scope to include multimodal fusion, effectively processing inputs from various sensors to enhance the overall perception capabilities.

Vision Transformers have brought about significant innovations in 3D object detection, with models like PETR \cite{liu2022petr, liu2023petrv2}, CrossDTR \cite{tseng2023crossdtr}, BEVFormer \cite{li2022bevformer, yang2023bevformer}, and UVTR \cite{li2022unifying} leading the way. PETR notably uses position embedding transformations to enhance image features with 3D coordinate information, offering a more detailed spatial understanding. CrossDTR integrates the strengths of DETR3D and PETR to create a unified framework for detection that is informed by cross-view analysis and depth guidance. BEVFormer utilizes a spatio-temporal Vision Transformer architecture, achieving unified BEV representations by integrating spatial and temporal data seamlessly. UVTR, on the other hand, specializes in depth inference, employing cross-modal interactions to form distinct voxel spaces, thus enabling an extensive multimodal analysis crucial for accurate 3D object detection.

The field of 3D segmentation in autonomous driving has seen significant improvements with the integration of Vision Transformers. Models like TPVFormer \cite{huang2023tri}, VoxFormer \cite{li2023voxformer}, and SurroundOcc \cite{wei2023surroundocc} are notable examples. TPVFormer reduces the computational load by converting volumes into BEV planes, maintaining high accuracy in semantic occupancy predictions. VoxFormer uses 2D images to create 3D voxel query proposals, enhancing segmentation through deformable cross-attention queries. SurroundOcc utilizes a distinctive approach to extract 3D BEV features from 2D images of varying views and scales, merging these features proficiently to map out densely occupied spaces.

Vision Transformer models have brought about transformative changes in 3D object tracking for autonomous vehicles. Models like MOTR \cite{zeng2022motr} and MUTR3D \cite{zhang2022mutr3d} have extended the capabilities of traditional tracking methods. MOTR, building on the DETR model, introduces a “track query” mechanism for modeling temporal variations across video sequences, avoiding reliance on conventional heuristics. MUTR3D introduces an innovative method that allows for concurrent detection and tracking. It employs associations across different cameras and frames to comprehend the three-dimensional state and appearance of objects over time, thereby greatly improving tracking accuracy and efficiency in autonomous driving systems.

\subsection{2D Perception Tasks}
In Autonomous Driving, tasks related to 2D perception include crucial functions like detecting lanes, segmenting various elements, and creating high-definition maps. These tasks focus on processing and understanding two-dimensional spatial data, a critical aspect of autonomous vehicle technology. Unlike 3D tasks, which deal with depth and volume, 2D tasks require a precise interpretation of flat surfaces and plane elements, crucial for the accurate navigation and safety of autonomous vehicles.

Lane detection is a primary area where Transformer models have been effectively utilized, categorized into two distinct groups. The first group includes models like BEVSegFormer \cite{peng2023bevsegformer}, which employs cross-attention mechanisms for multi-view 2D image feature extraction and CNN-based semantic segmentation for accurate lane marking detection. Another example, PersFormer \cite{chen2022persformer}, combines CNNs for 2D lane detection with Transformers for enhancing BEV features. The second group, featuring models like LSTR \cite{liu2021end} and CurveFormer \cite{bai2023curveformer}, focuses on directly generating road structures from 2D images. These models use Transformer queries to refine road markings and implement curve queries for effective lane line generation, demonstrating the versatility and precision of Transformers in lane detection tasks.

Beyond lane detection, Transformer models are increasingly applied to segmentation tasks within autonomous driving. TIiM \cite{saha2022translating} exemplifies this application with its sequence-to-sequence model that efficiently converts images and videos into overhead BEV maps, linking vertical scan lines in images to corresponding rays in maps for data-efficient and spatially aware processing. Panoptic SegFormer \cite{li2022panoptic} provides an all-encompassing approach to panoptic segmentation, integrating both semantic and instance segmentation. Utilizing a supervised mask decoder and a strategy for query decoupling, it improves the segmentation  efficiency. This model exemplifies the flexibility of Transformer architectures in handling intricate segmentation tasks.

In the realm of high-definition map generation, Transformer architectures like STSU \cite{can2021structured}, VectorMapNet \cite{liu2023vectormapnet}, and MapTR \cite{liao2022maptr} are bringing significant advancements. STSU treats lanes as directed graphs, focusing on learning Bezier control points and graph connectivity to convert front-view camera images into detailed BEV road structures. On the other hand, VectorMapNet leads the way in end-to-end vectorization of high-precision maps, utilizing sparse polyline primitives to model geometric shapes. MapTR offers an online framework for vectorized map generation, treating map elements as point sets and employing a hierarchical query embedding scheme. These models underscore the progress in merging multi-view features into a cohesive BEV perspective, crucial for creating accurate and detailed maps for autonomous driving.

\subsection{Prediction, Planning and Decision-Making Tasks}
Transformers are increasingly pivotal in autonomous driving, notably in prediction, planning, and decision-making. This progression marks a significant shift towards end-to-end deep neural network models that integrate the entire autonomous driving pipeline, encompassing perception, planning, and control into a unified system. This holistic approach reflects a substantial evolution from traditional models, indicating a move towards more comprehensive and integrated solutions in autonomous vehicle technology.

In trajectory and behavior prediction, Transformer-based models like VectorNet \cite{gao2020vectornet}, TNT \cite{zhao2021tnt}, DenseTNT \cite{gu2021densetnt}, mmTransformer \cite{liu2021multimodal}, and AgentFormer \cite{yuan2021agentformer} have addressed the limitations of standard CNN models, particularly in long-range interaction modeling and feature extraction. VectorNet enhances the depiction of spatial relationships by employing a hierarchical graph neural network, which is used for high-definition maps and agent trajectory representation. TNT and DenseTNT refine trajectory prediction, with DenseTNT introducing anchor-free prediction capabilities. The mmTransformer leverages a stacked architecture for simplified, multimodal motion prediction. AgentFormer uniquely allows direct inter-agent state influence over time, preserving crucial temporal and interactional information. WayFormer \cite{nayakanti2023wayformer} further addresses the complexities of static and dynamic data processing with its innovative fusion strategies, enhancing both efficiency and quality in data handling.

End-to-end models in autonomous driving have evolved significantly, particularly in planning and decision-making. TransFuser \cite{chitta2022transfuser, lai2023sensor} exemplifies this evolution with its use of multiple Transformer modules for comprehensive data processing and fusion. NEAT \cite{chitta2021neat} introduces a novel mapping function for BEV coordinates, compressing 2D image features into streamlined representations. Building upon this, InterFuser \cite{shao2023safety} proposes a unified architecture for multimodal sensor data fusion, enhancing safety and decision-making accuracy. MMFN \cite{zhang2022mmfn} expands the range of data types to include HD maps and radar, exploring diverse fusion techniques. STP3 \cite{hu2022st} and UniAD \cite{hu2023planning} further contribute to this field, with STP3 focusing on temporal data integration and UniAD reorganizing tasks for more effective planning. These models collectively mark a significant stride towards integrated, efficient, and safer autonomous driving systems, demonstrating the transformative impact of Transformer technology in this domain.

\section{Open Challenges and Future Directions}
Vision Transformers offer promise for Autonomous Driving but face hurdles like data collection, safety, and interpretability. While they excel in perception and prediction, trends like multimodal fusion and explainability are emerging. Future focus areas include real-time processing optimization and end-to-end model development, necessitating continued research to overcome these challenges.

\textbf{Challenges in Implementing Transformer Models.} Transformers, evolving from their initial focus on 3D obstacle perception to a range of perception tasks in autonomous driving, encounter new challenges as they move towards integrating multimodal fusion. This transition necessitates multimodal models as well as efficient wireless connection \cite{ipark, park} to enhance efficiency gains, crucial for advanced autonomous driving. However, it also brings complexities in training and requires advancements in algorithms and system integration.

\textbf{Hardware Acceleration and Model Complexity.} As Transformer models grow in complexity, they demand innovative hardware acceleration solutions for efficient deployment. Integrating various models into an end-to-end system poses challenges in hardware optimization, especially with operators like Deformable Attention that complicate parallel processing. This necessitates specialized hardware designs to meet the diverse requirements of these advanced models.

\textbf{Future Directions: Algorithm and Hardware Advancements.} Future advancements in Transformer models for autonomous driving will focus on algorithm enhancements and hardware innovations. Key areas include mixed-precision quantization for balancing computational demands and interpretability techniques like attention-based saliency maps. These developments aim to improve model compression and offer insights into decision-making processes, building trust in autonomous systems.

\textbf{Enhancing Model Efficiency and Interpretability.} Efficiency and interpretability are pivotal for the future of Transformer models in autonomous driving. The need for models that process multi-view data effectively and offer improved generalization while being optimized for performance is critical. Developing interpretable models with techniques to visually highlight crucial data will enhance system reliability and user trust in autonomous driving technology.

\section{Conclusion}
This paper has offered a comprehensive survey of Transformer models, especially Vision Transformers, in autonomous driving (AD), demonstrating their significance beyond traditional Convolutional Neural Networks (CNNs) and Recurrent Neural Networks (RNNs). We explored their foundational architecture, attention-based processing advantages in natural language processing and computer vision, and their superior performance in various AD tasks, including 3D object detection, 2D lane detection, and advanced scene analysis. Additionally, we highlighted challenges, trends, and future perspectives of Vision Transformers in AD, aiming to spur further interest and research in this dynamic field. The potential of Vision Transformers in transforming AD, with their nuanced data processing capabilities, promises exciting advancements in vehicular technologies.

\nocite{langley00}

\bibliography{icml2023/mypaper}
\bibliographystyle{ieeetr}


\end{document}